\documentclass[runningheads]{llncs}

\usepackage[T1]{fontenc}
\usepackage[utf8]{inputenc}
\usepackage{microtype}    
\usepackage{amsmath}      
\usepackage{amssymb}      
\usepackage{graphicx}     
\usepackage{booktabs}     
\usepackage[colorlinks=true, linkcolor=blue, citecolor=blue, urlcolor=blue]{hyperref}
\usepackage[inline]{enumitem}

\usepackage{tikz}
\usepackage{pgfplots}
\usepgfplotslibrary{groupplots}
\pgfplotsset{compat=1.18}
\usetikzlibrary{decorations.text} 
\usetikzlibrary{arrows.meta} 
\usetikzlibrary{positioning} 
\usepackage[markup=underlined]{changes} 
\usepackage{xcolor}
\usepackage{amssymb}
\usepackage{multirow}
\usepackage{bbding}

\usepackage{bibunits}
\defaultbibliographystyle{splncs04}

\definechangesauthor[name={Elena Casiraghi}, color=red]{EC}
\definechangesauthor[name={Pietro Miotto}, color=purple]{PM}
\definechangesauthor[name={Mauricio Soto-Gomez}, color=violet]{MS}
\definechangesauthor[name={Giorgio Valentini}, color=blue]{GV}
\definechangesauthor[name={Lucia Mellini}, color=orange]{LM}
 
\begin{document}

%\begin{bibunit}

\title{Hyperbolic Graph Representation Learning for Differential Diagnosis on Biomedical Knowledge Graphs}
\titlerunning{Hyperbolic Bio-KGRL for Differential Diagnosis}

\author{
Pietro Miotto\inst{1,2}\orcidID{0009-0002-2973-0050}\Envelope\and
Lucia Mellini\inst{1}\orcidID{0009-0007-4186-8745}\and
Tommaso Marzi\inst{2}\orcidID{0000-0002-8232-9621}\and
Cesare Alippi\inst{2,3}\orcidID{0000-0003-3819-0025}\and
Elena Casiraghi\inst{1,4}\orcidID{0000-0003-2024-7572}\and
Alberto Paccanaro\inst{5}\orcidID{0000-0001-8059-1346}\and 
Giorgio Valentini\inst{1}\orcidID{0000-0002-5694-3919}\and
Mauricio Soto-Gomez\inst{1}\orcidID{0000-0001-5977-9467}
}

\authorrunning{P. Miotto et al.}

\institute{
Department of Computer Science, Università degli Studi di Milano, Milan, Italy \and 
Università della Svizzera italiana (USI), Lugano, Switzerland
% \\
%  \email{miottpi@usi.ch}\\
 \and
Politecnico di Milano, Milan, Italy \and
Department of Computer Science, Aalto University, Espoo, Finland \and
School of Applied Mathematics (EMAp), FGV, Rio de Janeiro, Brazil\\
%\vspace{0.1cm}
 Correspondence to: \email{miottpi@usi.ch}
}

\maketitle

\begin{abstract}
Biomedical knowledge graphs combine ontology-derived hierarchies with transversal associations among heterogeneous entities such as phenotypes, diseases, genes, proteins, and patients. This hybrid structure raises the question of whether hyperbolic embeddings, which naturally capture tree-like organization, remain useful beyond purely hierarchical graphs. We present a preliminary study of hyperbolic graph representation learning for Mendelian-disease differential diagnosis on a patient-integrated biomedical graph. Experiments on isolated ontology subgraphs show that hyperbolic models achieve strong performance in substantially lower dimensions than Euclidean baselines. We then evaluate the models on a link-prediction task that ranks candidate diseases for each patient. Results suggest that hyperbolic embeddings can exploit biomedical hierarchical structure while supporting diagnostic reasoning over heterogeneous patient-level graphs.

\keywords{Knowledge Graphs Representation Learning \and Hyperbolic Geometry \and Differential Diagnosis \and GNNs \and Biomedical Graphs}

\end{abstract}

%\vspace{0.5cm}

\section{Introduction}
Biomedical knowledge is inherently structured. Not rarely, a large fraction of it is organized through ontologies, which represent concepts and their dependencies by means of hierarchical relations. Disease taxonomies \cite{Vasilevsky2022}, phenotype descriptions \cite{HumanPhenotypeOntology}, gene functions \cite{Ashburner2000}, and many other biomedical resources are represented in this way, typically as Directed Acyclic Graphs (DAGs). When such ontologies are integrated with other biomedical entities and associations into Knowledge Graphs (KGs), the resulting structures preserve a strong hierarchical backbone while also incorporating additional non-hierarchical relations across domains \cite{torgano2025rna}.

Once biomedical knowledge is represented as a graph, Graph Representation Learning (GRL) methods map nodes and relations into a continuous latent space so that the geometry of the embeddings reflects the topology of the original graph as faithfully as possible, given a suitably controlled space. In most existing approaches, this latent space is assumed to be Euclidean~\cite{torgano2025rna}. However, this choice is neither appropriate in general nor always compatible with the intrinsic underlying hierarchical structure of biomedical KGs. 
In such settings, the number of nodes typically grows exponentially with the hierarchy-depth, whereas Euclidean volume grows only polynomially with radius. As a result, hierarchical and tree-like DAGs are poorly represented in flat spaces, where embeddings inevitably incur distortion and loss of information \cite{bendavid2002limitations}. A hyperbolic space, by contrast, has constant negative curvature and volume that grows exponentially with distance from the origin. This geometric match allows hierarchical structures to be embedded with substantially lower distortion, often even in low dimensions, as shown by seminal work on Poincar\'e embeddings and later hyperbolic representation learning studies \cite{chami2021representation,nickel2017poincare}. Nevertheless, real biomedical KGs also contain transversal (non-hierarchical) connections linking different semantic domains, such as phenotype--disease associations, gene--disease links, protein interactions, and patient similarities. These edges do not necessarily follow a purely hierarchical organization and may instead exhibit structural patterns that are better captured by Euclidean or mixed geometries. An important open question is whether hyperbolic embeddings retain their well-established advantage also on heterogeneous biomedical graphs, or whether partially hierarchical environments call for hybrid representations instead.

In this work, we aim to address two complementary questions.
(i) How do Euclidean and hyperbolic methods compare in representing ontology-specific subgraphs? In particular, hyperbolic methods are expected to be most advantageous when hierarchical structure dominates; however, they may fail to fully capture the topology of ontology-based KGs when hierarchical DAGs are interconnected with non-hierarchical relationships. Recent analysis support this claim showing that the benefit of hyperbolic GNNs emerges on geometry-aligned tasks such as link prediction over strongly tree-like graphs (low $\delta$-hyperbolicity), and becomes negligible otherwise~\cite{naddeo2026microscope}
% Recent analysis support this claim showing that the benefit of hyperbolic GNNs emerges on geometry-aligned tasks such as link prediction over strongly tree-like (low $\delta$-hyperbolicity) graphs, and weakens as non-hierarchical structure is introduced 
(ii) Do the geometric biases of hyperbolic models, including Hyperbolic Graph Neural Networks such as HGCNs~\cite{chami2019HGCN} and shallow hyperbolic models such as AttH~\cite{AttH}, translate into useful predictive behavior in a clinically motivated Mendelian-disease ranking task?

To investigate these questions systematically, we construct a biomedical graph integrating disease and phenotype ontologies with the phenotypic profiles of patients affected by Mendelian diseases. This graph enables us to formulate Mendelian-disease differential diagnosis as a link prediction problem over edges connecting patient nodes to disease nodes. In this setting, candidate diseases for a patient are ranked according to the predicted probability of the corresponding patient--disease links.

This patient--disease ranking problem provides a focused and clinically relevant test bed for evaluating whether differences between Euclidean and hyperbolic representations translate into useful predictive behavior in biomedical graphs. Mendelian diseases constitute a clinically important subset of rare diseases and are typically caused by pathogenic variants in single genes. Rare disease diagnosis is particularly suitable for this evaluation, as it remains one of the most challenging problems in modern medicine. Although each rare disease affects only a limited number of individuals, more than 10{,}000 rare diseases collectively impact about 5\% of the global population~\cite{Haendel20}, making their overall burden substantial.

\section{Background and Methods}
\label{sec:background}

\paragraph{\bf Graph Representation Learning}
%Graph Representation Learning 
(GRL) aims to map nodes and edges from a discrete graph domain into a 
low-dimensional continuous latent space through an encoder function $\text{ENC}$. 
These representations are defined so that their geometric proximity, measured by a decoder function 
$\text{DEC}$, reflects the structural and semantic properties of the graph~\cite{hamilton2020graph}. 
The learned representations can then be used for downstream tasks such as link prediction, 
entity classification, or graph-level inference. The factual knowledge encoded in Knowledge Graphs (KGs) is intrinsically multi-relational, expressed as triples \textit{(head entity, relation, tail entity)} $\in \mathcal{V}\times\mathcal{R}\times\mathcal{V}$, where $\mathcal{V}$ and $\mathcal{R}$ denote the sets of entities and relations, respectively.
%
%In this context, a useful conceptual and historical distinction is between \emph{shallow} and \emph{non-shallow} GRL methods.  
Knowledge Graph Embedding (KGE) methods can be broadly classified along two dimensions: (1) the geometry of the embedding space, such as Euclidean or hyperbolic; and (2) the mechanism by which representations are obtained, distinguishing \emph{shallow} from \emph{non-shallow} methods~\cite{hamilton2020graph}.\\

% \paragraph{Shallow Euclidean GRLs.}
% Shallow Euclidean models \cite{DeepWalk,TransE,DistMult} dominate GRL due to their higher scalabitily. Embeddings are computed as free parameters to be optimized. These approaches are therefore inherently \emph{transductive}: they learn representations only for entities observed during training, and extending them to unseen nodes generally requires retraining. 
% Among them, translational methods interpret relations as vectorial displacements or relation-specific geometric transformations \cite{TransE}.%,RotatE}.
% The canonical and simplest example is \textbf{TransE} \cite{TransE}, which models a relation, $r$, as a translation from head, $\mathbf{e}_h$, to tail, $\mathbf{e}_t$:
% \begin{equation}
%     \text{DEC}(\mathbf{e}_h,r,\mathbf{e}_t)= -\|\mathbf{e}_h+\mathbf{r}_r-\mathbf{e}_t\|.
% \end{equation}
% or by its extensions \cite{hamilton2020graph} that capture highly complex relational patterns by applying learnable transformations to the head and tail embeddings. %(see Appendix \ref{app:TransE_extension}). 
% The simple formulation driving translational models established the influential idea that KG relations can be represented as geometric operators in latent space. 

In \emph{shallow} methods, the encoder is simply an embedding lookup table, mapping each entity 
and relation to a vector via their identifiers.
The embedding vectors are therefore directly the free parameters of the optimization process, whose objective is to find values that best preserve some notion of 
structural proximity as measured by $\text{DEC}$.
Shallow methods are inherently \emph{transductive}: embeddings are tied to entities seen during training, and incorporating unseen entities requires either full retraining or ad-hoc approximations.

Notably, \emph{translational models} interpret relations as geometric transformations in the latent space.
The foundational example is \textbf{TransE}~\cite{TransE}, which models a relation $\mathbf{r}_r$ as a translation from the head embedding $\mathbf{e}_h$ to the tail embedding $\mathbf{e}_t$. This is done by minimizing:
 \begin{equation}
     \text{DEC}(\mathbf{e}_h,\mathbf{r}_r,\mathbf{e}_t)= \|\mathbf{e}_h+\mathbf{r}_r-\mathbf{e}_t\|
\end{equation}
for existing triples while maximizing it for non-existing ones.
Extensions of this framework introduce increasingly expressive geometric operators as entity transformations
~\cite{hamilton2020graph},
relation-specific hyperplanes~\cite{TransH},
% separate entity and relation spaces~\cite{TransR}, 
or complex-valued transformations~\cite{RotatE}.\\

\emph{Non-shallow} methods address the transductive limitation by recasting the embedding task from an optimization problem into a function learning problem.
They learn parameterized functions that compute representations from local neighborhood structure and entity features.
These functions are typically implemented as Graph Neural Networks (GNNs) \cite{GCN,GAT}, which aggregate information from local neighborhoods so that topology acts as an inductive bias constraining information flow.
In principle, these methods can generalize to previously unseen nodes provided their features and connectivity are available \cite{hamilton2020graph}.
In the KG setting, this family includes relation-aware, convolutional, and attention-based architectures~\cite{RGCN}, %such as R-GCN, ConvE, ConvKB \cite{RGCN,CONVE,ConvKB},
as well as more recent path-based~\cite{NBFnet} and tokenization approaches.
% such as NBFNet, AStarNet, NodePiece, and KGT5 \cite{NBFnet,AStarNet,NodePiece,KGT5}. 
% 
% These methods offer greater flexibility and potential inductive generalization, but on dense multi-relational graphs they can be computationally demanding, may suffer from over-smoothing, and still inherit the geometric mismatch of Euclidean space when the data are strongly hierarchical.
Despite the flexibility of non-shallow Euclidean methods, they introduce greater architectural complexity, can be computationally demanding on dense multi-relational graphs, and are susceptible to over-smoothing.
Furthermore, on strongly hierarchical graphs exhibiting scale-free properties, the message-passing procedure can cause neighborhood explosion around high-degree nodes.%, and the use of \highlight[id=PM, comment= correct but this is solved by tangent space aggregation]{Euclidean aggregation introduces a geometric mismatch.}

%\vspace{-0.2cm}
\paragraph{\bf Hyperbolic Geometry Models.}
Many KGs exhibit scale-free, tree-like topologies in which the number of entities grows exponentially with relational depth,  a latent structure that Euclidean spaces cannot faithfully represent in low dimensions.
\emph{Hyperbolic spaces}, by contrast, have constant negative curvature, which endows them with an exponentially expanding volume that naturally accommodates hierarchical structure. This allows for tree-like graph representations with low distortion even in very low dimensions~\cite{nickel2017poincare}.

%\paragraph{Hyperbolic Learning.} 
From a computational perspective, hyperbolic learning is usually implemented by exploiting the fact that hyperbolic space is a Riemannian manifold, and therefore admits a Euclidean tangent space at every point. 
Optimization and linear operations can thus be performed in the tangent space at the origin of the hyperbolic manifold, %where standard Euclidean calculus applies, 
and then mapped back to the hyperbolic manifold. 
The interface between the curved manifold and its local flat approximation is achieved through logarithmic and exponential maps.
Since ordinary (Euclidean) vector addition is not geometry-preserving in hyperbolic space, it is replaced by M\"obius addition, the hyperbolic analog to the vector addition in Euclidean vector spaces. 
The logarithmic and exponential maps, together with M\"obius addition, make it possible to generalize Euclidean models to hyperbolic latent spaces while retaining computational efficiency. 
%Detailed definitions and formulas are reported in Appendix~\ref{app:poincare_ball}.
% \paragraph{Hyperbolic Geometry Models: the Poincar\'e Ball.}
% Among the models of hyperbolic space, the Poincar\'e ball is one of the most widely used because it provides an intuitive and conformal representation of negatively curved geometry \cite{nickel2017poincare}. In this model, points lie inside an open Euclidean ball, while distances increase rapidly near the boundary, allowing more general concepts to remain close to the origin and increasingly specific concepts to be placed progressively farther away. This makes the Poincar\'e ball particularly suitable for representing hierarchical structures.
%\smallskip

\paragraph{Shallow hyperbolic models}\label{pg:shallow_Hmodels}
embed entities directly in a hyperbolic manifold, most commonly the Poincar\'e ball or the hyperboloid model, and define scoring functions using hyperbolic distance.
Poincar\'e embeddings~\cite{nickel2017poincare} demonstrate the representational efficiency of this approach on hierarchical data. 
MuRP~\cite{MuRP} extends this idea to multi-relational KGs, generalizing translational models to the Poincar\'e ball.

Among shallow hyperbolic KG embedding methods, \textbf{AttH} \cite{AttH} is particularly relevant because it extends the translational paradigm while enriching relation modeling through relation-specific isometries and attention. A key contribution of AttH is the use of a relation-specific curvature $c_r$, which allows different relations to be represented in hyperbolic spaces with different effective hierarchical capacities. For a triple $(h,r,t)$ in the KG, AttH first learns two relation-specific transformations of the head embedding $\mathbf{e}_h^E$ in the tangent Euclidean space at the origin of the hyperbolic manifold $o$. 
Specifically, it defines a rotation and a reflection, both implemented through matrix Givens transformations:
\begin{equation}
\mathbf{q}_{\mathrm{Rot}}^E=\mathrm{Rot}(\Theta_r)\mathbf{e}_h^E,
\qquad
\mathbf{q}_{\mathrm{Ref}}^E=\mathrm{Ref}(\Phi_r)\mathbf{e}_h^E.
\end{equation}
with $\mathrm{Rot}(\Theta_r)$ and $\mathrm{Ref}(\Phi_r)$ being the rotation and reflection operators, where $\Theta_r$ and $\Phi_r$ are the respective learned parameters.  
These two transformed views are then combined through a simple attention mechanism, i.e., a weighted average in the tangent Euclidean space between $\mathbf{q}_{\mathrm{Rot}}^E$ and $\mathbf{q}_{\mathrm{Ref}}^E$. 
Having learned the relation-specific attention vector $\mathbf{a}_r = [\alpha_{\mathrm{Rot}}, \alpha_{\mathrm{Ref}}]$ the combined representation is then mapped to the hyperbolic space:
\begin{equation}
    \begin{cases}
         \mathrm{Att}^E(\mathbf{q}_{\mathrm{Rot}}^E,\mathbf{q}_{\mathrm{Ref}}^E;\mathbf{a}_r)
         =
        \alpha_{\mathrm{Rot}}\mathbf{q}_{\mathrm{Rot}}^E+
        \alpha_{\mathrm{Ref}}\mathbf{q}_{\mathrm{Ref}}^E
        \\
         \mathrm{Att}^H(\mathbf{q}_{\mathrm{Rot}}^E,\mathbf{q}_{\mathrm{Ref}}^E;\mathbf{a}_r)
        =
        \exp_o^{c_r}\!\left(        \mathrm{Att}^E(\mathbf{q}_{\mathrm{Rot}}^E,\mathbf{q}_{\mathrm{Ref}}^E;\mathbf{a}_r)
        \right) 
    \end{cases}.
\end{equation}

A translation is then performed on the manifold with a relation-specific embedding $\mathbf{r}_r^H$ through M\"obius addition resulting in a so-called query:
\begin{equation}
\mathbf{q}^H(h,r)=
\mathrm{Att}(\mathbf{q}_{\mathrm{Rot}}^E,\mathbf{q}_{\mathrm{Ref}}^E;\mathbf{a}_r)
\oplus^{c_r}\mathbf{r}_r^H.
\end{equation}

Similarly to the Euclidean case, the query is expected to lie near the embedding of the relation tail in the hyperbolic manifold $\mathbf{e}_t^H$, obtained by projecting the Euclidean tail embedding $\mathbf{e}_t^E$ into the Poincar\'e ball via the exponential map at the origin $o$:
$
\mathbf{e}_t^H = \exp_o^{c_r}(\mathbf{e}_t^E).
$
The score for the triple, $s(h,r,t)$ is then computed as the negative squared Poincar\'e distance between the query $\mathbf{q}^H(h,r)$ and the tail embedding $\mathbf{e}_t^H$:
\begin{equation} \label{eq:AttH_score}
    s(h,r,t)= -\big(d_{\mathbb{B}}^{c_r}(\mathbf{q}^H(h,r),\mathbf{e}_t^H)\big)^2+b_h+b_t,
\end{equation}
where $b_h$ and $b_t$ are entity-specific biases applied to the head and tail respectively.

AttH is typically trained with a cross-entropy objective and uniform negative sampling. For each positive triple $(h,r,t)$, negative examples are generated by corrupting the tail with entities sampled uniformly from the node set $\mathcal{V}$:
\begin{equation}
\label{eq:ATTH_loss}
\mathcal{L} =
\sum_{t' \sim \mathcal{U}(\mathcal{V})}
\log\!\left(1+\exp\!\big(y_{t'}\, s(h,r,t')\big)\right),
\qquad
y_{t'}=
\begin{cases}
-1 & \text{if } t'=t\\
\phantom{-}1 & \text{otherwise.}
\end{cases}
\end{equation}
where $\mathcal{U}(\mathcal{V})$ denotes uniform sampling over the entity set. This objective encourages the translated head embedding to move closer to the true tail embedding for positive triples, while pushing it away from corrupted tails for negative triples.

The Euclidean ablation of AttH, \textbf{AttE} \cite{AttH}, removes curvature and replaces hyperbolic distance and M\"obius addition with their Euclidean counterparts. In this way, the query is defined as $\mathbf{q}(h,r)^E = \mathrm{Att}^E(\mathbf{q}_{\mathrm{Rot}}^E,\mathbf{q}_{\mathrm{Ref}}^E;\mathbf{a}_r) + r_r^E $ and the scoring function as $s(h,r,t) = -\|\mathbf{q}(h,r)^E - \mathbf{e}_t^E\|_2^2 + b_h + b_t$. 
There is no curvature parameter.

\paragraph{Non-shallow hyperbolic models: HGNN and HGCN.}
Hyperbolic geometry has also been extended to non-shallow models, where node embeddings are computed through message passing rather than learned as free parameters. In \textbf{HGNN}~\cite{HGNN}, this is achieved by generalizing GNN updates to Riemannian manifolds: representations of node neighborhoods are mapped to a tangent space centered at the origin, aggregated there with standard linear operations, and then projected back to the manifold through logarithmic and exponential maps. This provides the basic manifold-aware message-passing template for hyperbolic GNNs.

Building on this idea, \textbf{HGCN}~\cite{chami2019HGCN} makes the framework more expressive and practically robust in three ways: 
\begin{enumerate*}[label=(\roman*)]
    \item it explicitly maps Euclidean input features into hyperbolic space;
\item it uses layer-specific MLP-learned attention to aggregate neighborhood features in the tangent space centered at the target node, more faithfully reflecting local geometry; and
    \item it introduces \emph{layer-wise trainable curvature}, allowing different layers to operate in hyperbolic spaces of different curvature.
\end{enumerate*}
As a result, a HGCN layer can be viewed as a three-step pipeline: hyperbolic feature transformation, geometry-aware neighborhood aggregation, and curvature-aware nonlinearity.
Conceptually, HGNN provides the general recipe for hyperbolic message passing, whereas HGCN refines it into an inductive architecture better suited to hierarchical and scale-free graphs. %by combining Euclidean-to-hyperbolic feature mapping, local hyperbolic attention, and trainable curvature across different HGCN layers. 

\paragraph{\bf Hyperbolic, Mixed-Geometry, and Biomedical KG Models.}

Hyperbolic \cite{AttH,chami2019HGCN,HGNN} and mixed-geometry models \cite{M2GNN} are increasingly relevant in biomedicine, where ontologies, protein interaction networks, and biomedical knowledge graphs exhibit strong latent non-Euclidean structure. Interest in this direction expanded after \cite{AlanisLobato2018} showed that the human protein interaction network displays latent hyperbolic geometry. Subsequent work explored Poincar\'e-based embeddings of the Gene Ontology and gene annotations \cite{GeOKG}, hierarchical drug representations \cite{HyperDrug}, and drug-target association prediction \cite{HyperMF}. More recent approaches extended hyperbolic learning to broader biomedical KGs through mixed-curvature spaces \cite{HEM} and product manifolds \cite{ProductManifoldBIO} for gene-disease and protein interaction modeling.

Despite this progress, most studies remain focused on molecular-scale tasks such as ontology embedding, protein interactions, and drug-target prediction. The application of non-Euclidean geometry to patient-level phenotypic reasoning and differential diagnosis remains largely unexplored, even though phenotype and disease ontologies such as HPO and MONDO are strongly hierarchical. Rare disease diagnosis has instead been dominated by semantic similarity and ontology-based methods \cite{Phenomizer,Phen2Disease,Exomiser}. More recently, SHEPHERD \cite{SHEPHERD} introduced a Euclidean GNN over a patient-augmented biomedical KG for phenotype-driven diagnosis, while LLM-based pipelines have begun to exploit biomedical graphs for diagnostic reasoning \cite{Chen2024RareBenchSpecialists}. However, existing biomedical KGs such as PrimeKG \cite{chandak2023primekg} still lack direct patient-disease connectivity, and the role of latent geometry in patient-integrated KGs remains poorly understood.

\smallskip
Overall, the evolution from Euclidean embeddings to hyperbolic and neural message-passing models motivates our study of whether negatively curved latent spaces better capture the hierarchical structure of biomedical KGs while balancing expressive relation modeling and inductive graph learning.

\section{Experimental Results}
\begin{figure}[tbph]
\centering
% \resizebox{width}{height}{content}. Using "!" keeps the aspect ratio proportional.
\resizebox{0.85\linewidth}{!}{%
\begin{tikzpicture}[scale=.9]
\usetikzlibrary{decorations.text}

\begin{scope}[xshift=1cm]
% Adjusted bounding box width from 8 to 11 to include the Male/Female nodes
%\useasboundingbox (-2, 2.5) rectangle (10, 11);

% Nodes 
\node[circle, draw, fill=none, 
    minimum  size=1.22cm%2.836cm
, inner sep=0pt] (Disease) at (2.5,-1.1) {Disease};
\node[circle, draw, fill=none, 
    minimum size=1.15cm%2.119cm
, inner sep=0pt] (Person) at (6,0) {Person};
% \node[circle, draw, fill=none, minimum size=1.000cm, inner sep=0pt] (Article) at (-2,11.5) {Article};
\node[circle, draw, fill=none, 
    minimum size=1.6cm%2.655cm
    , inner sep=0pt] (Phenotype) at (-2,0) {Phenotype};

% Male and Female nodes 
\node[circle, draw, dotted, fill=none, minimum size=1.2cm, inner sep=0pt] (Male) at (10,0.5) {Male};
\node[circle, draw, dotted, fill=none, minimum size=1.2cm, inner sep=0pt] (Female) at (9.5,-0.9) {Female};

% Invisible nodes for dotted edges
% \node[coordinate] (Invisible1) at (3,1) {};
% \node[coordinate] (Invisible2) at (-1,2) {};

%self edges -----------

\draw[->, line width=1.5pt, opacity=0.5158] (Disease) to[out=130, in=50, looseness=4] (Disease);
\path[decorate, decoration={text along path, text align=center, raise=2pt, text={Subclassof}}] (Disease) to[out=130, in=50, looseness=4] (Disease);

\draw[->, line width=1.5pt, opacity=0.3981] (Phenotype) to[out=225, in=135, looseness=4] (Phenotype);
\path[decorate, decoration={text along path, text align=center, raise=2pt, text={Subclassof}}] (Phenotype) to[out=225, in=135, looseness=4] (Phenotype);

\draw[->, line width=1.5pt, opacity=0.1000] (Person) to[bend left=20] (Male);
\path[decorate, decoration={text along path, text align=center, raise=2pt, text={Subclassof}}] (Person) to[bend left=20] (Male);

\draw[->, line width=1.5pt, opacity=0.1000] (Person) to[bend right=20] (Female);
\path[decorate, decoration={text along path, text align=center, raise=2pt, text={Subclassof}}] (Person) to[bend right=20] (Female);
%--------------------

% \draw[->, line width=1.5pt, opacity=0.5871] (Phenotype) to[bend right=10] (Person);
% \path[decorate, decoration={text along path, text align=center, raise=2pt, text={ Phenotype Of}}] (Phenotype) to[bend right=10] (Person);

\draw[->, line width=1.5pt, opacity=0.5871] (Person) to[bend right=20] (Phenotype);
\path[decorate, decoration={text along path, text align=center, raise=2pt, text={Has Phenotype}}] (Phenotype) to[bend left=20] (Person);

% \draw[->, line width=1.5pt, opacity=0.1463] (Article) to[bend right=20] (Person);
% \path[decorate, decoration={text along path, text align=center, raise=2pt, text={Part Of}}] (Article) to[bend right=20] (Person);

% \draw[->, line width=1.5pt, opacity=0.1463] (Person) to[bend right=20] (Article);
% \path[decorate, decoration={text along path, text align=center, raise=2pt, text={Has Part}}] (Article) to[bend left=20] (Person);

% \draw[->, line width=1.5pt, opacity=1.0000] (Phenotype) to[bend right=20] (Disease);
% \path[decorate, decoration={text along path, text align=center, raise=2pt, text={Phenotype of}}] (Phenotype) to[bend right=20] (Disease);

\draw[->, line width=1.5pt, opacity=1.0000] (Disease) to[bend left=30] (Phenotype);
\path[decorate, decoration={text along path, text align=center, raise=2pt, text={Has Phenotype}}] (Phenotype) to[bend right=30] (Disease);

\draw[->, line width=1.5pt, opacity=0.1463] (Person) to[bend left=30] (Disease);
\path[decorate, decoration={text along path, text align=center, raise=2pt, text={Has Disease}}] (Disease) to[bend right=30] (Person);

\end{scope}
\end{tikzpicture}%
}
\caption{\emph{Schema of the patient-integrated knowledge graph}. Solid-bordered nodes represent super-nodes that aggregate instances of the same class and summarize their interaction patterns. Dashed-bordered nodes represent attribute-specific subclasses used in our graph model to encode categorical attributes. Here, Male and Female, connected to Person by SubClassOf relations, encode the possible values of the sex attribute. Super-edge opacity indicates the relative frequency of the corresponding predicate in the dataset, with darker edges denoting higher frequencies.}
\label{fig:kg_hypergraph_integration}
\end{figure}
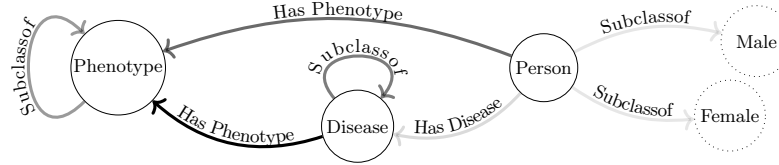

\paragraph{\textbf{KG construction.}}
We conduct experiments on a patient-integrated graph, constructed using the \emph{PheKnowLator} (\emph{Phenotype Knowledge Translator}) framework~\cite{callahan2020PheKnowlator}. The graph integrates disease ontologies with phenotypic information derived from patient-level clinical observations. 
In particular, the graph contains 28,811 nodes and 178,320 edges. The nodes belong to three node types: \texttt{Person} nodes, whose phenotypic information, i.e., symptoms, was gathered from the Phenopacket Store~\cite{PhenopacketStore}; \texttt{Phenotype} nodes, corresponding to Human Phenotype Ontology (HPO \cite{HumanPhenotypeOntology}) terms; \texttt{Disease} nodes derived from the Disease Ontology (DOID)~\cite{Schriml2012} and the Mondo Disease Ontology (MONDO)~\cite{Vasilevsky2022}. 
These nodes are connected through heterogeneous relations, encoding transversal patient--phenotype annotations, and patient--disease annotations, as well as hierarchical relationships within the HPO and disease ontologies.
Note that, due to the computational complexity of the HGCN framework~\cite{chami2019HGCN}, \texttt{Disease} nodes were restricted to those associated with the patient cohort, together with their ancestors. Conversely, \texttt{Phenotype} nodes were retained to preserve patient-level clinical descriptions and phenotype hierarchy. 
The resulting metagraph schema is shown in Figure~\ref{fig:kg_hypergraph_integration}.  
This graph enables graph representation learning (GRL) methods to exploit both ontology-derived hierarchical structure and cross-domain biomedical associations for differential diagnosis.
%\vspace{-1cm}
%\comment[id=PM]{despite the great idea of Lucia to interpret opacity as node frequency, which should be added on data representation books, I think that in order to avoid readers' complaints and confusion, we should not confuse the reader with logarithmic scaling and focus on the design primitives used. Note i used super-nodes instead of hyper-nodes because the term "hyper" could make reader think we are dealing with a hyper-graph}

%Schema of the proposed patient-integrated KG. Dashed borders indicate specific and peculiar node instances assigning the sex attribute to each node, solid borders represent super-nodes that summarize how instances of a given class interact. Edges are super-edges where opacity indicates the relative frequency of the predicates within the dataset (the darker the more frequent).

%}{Schema of the patient-integrated knowledge graph. Hyper-node size is scaled logarithmically by node frequency. Hyper-edge opacity is scaled logarithmically by predicate frequency.}}

%
%\vspace{-1cm}
\subsection{Binary Link Prediction on Isolated Hierarchies}

Following the standard protocol~\cite{nickel2017poincare}, within each ontology subgraph, we extract the \texttt{SubclassOf}-induced subtree and add its transitive closure, connecting every node to each of its ancestors through a direct \texttt{SubclassOf} arc. The models are trained on a standard binary link prediction task, i.e., to predict link existence against negative samples. %}{In particular, within each subgraph, we trained the models on a standard binary link prediction task, i.e., to predict link existence against negative samples, taking into account the transitive closure of the \texttt{SubclassOf}-induced subtrees}. 
%\replaced[id=PM, comment={being more explicit also here}]{
The edges of this enriched SubclassOf hierarchy are then randomly partitioned with an 80\%--10\%--10\% split into training, validation, and test sets. For each positive edge, one negative edge is sampled uniformly at random from the complement graph to yield negative sets balanced 1:1 with posistives.%}{For training, validation, and testing, we created balanced training, validation, and test graphs using an 80\%--10\%--10\% split of the positive edges, with an equal number of negative edges obtained by corrupting each positive tuple in the training set.}
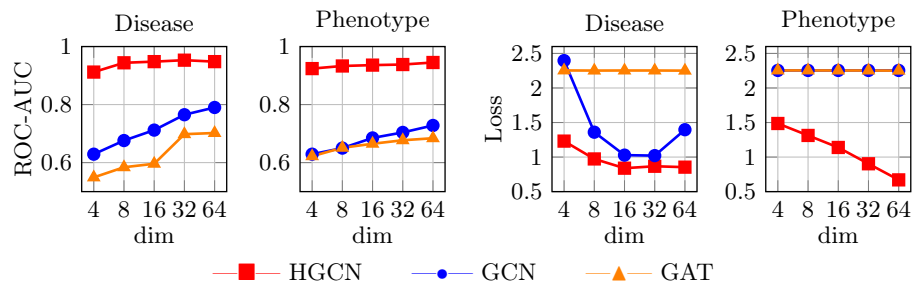
\begin{figure}[tbp]
    \centering
    % --- DISEASE (ROC-AUC) ---
    \begin{minipage}[b]{0.24\textwidth}
        \centering
        \begin{tikzpicture}
            \begin{axis}[
                axis line style={line width=0.5pt},
                width=3.5cm, 
                height=3.5cm,         
                title={Disease},
                title style={yshift=-1ex},
                ylabel={ROC-AUC},
                xlabel={dim},
                ylabel shift={-3pt},
                xlabel shift={-3pt},
                xmode=log,
                log basis x={2},
                xtick={4, 8, 16, 32, 64},
                xticklabels={4, 8, 16, 32, 64},
                tick label style={font=\footnotesize},
                ymin=0.5, ymax=1.0,
                grid=both,
                line width=1pt,
            ]
            \addplot[color=red, mark=square*] coordinates {(4,0.912) (8,0.944) (16,0.948) (32,0.953) (64,0.948)};
            \addplot[color=blue, mark=*] coordinates {(4,0.629) (8,0.676) (16,0.712) (32,0.765) (64,0.790)};
            \addplot[color=orange, mark=triangle*] coordinates {(4,0.549) (8,0.584) (16,0.596) (32,0.698) (64,0.702)};
            \end{axis}
        \end{tikzpicture}
    \end{minipage}\hfill
    % --- PHENOTYPE (ROC-AUC) ---
    \begin{minipage}[b]{0.25\textwidth}
        \centering
        \begin{tikzpicture}
            \begin{axis}[
                axis line style={line width=0.5pt},
                width=3.5cm,
                height=3.5cm,
                title={Phenotype},
                xlabel={dim},
                xlabel shift={-3pt},
                title style={yshift=-1ex},
                xmode=log,
                log basis x={2},
                xtick={4, 8, 16, 32, 64},
                xticklabels={4, 8, 16, 32, 64},
                tick label style={font=\footnotesize},
                ymin=0.5, ymax=1.0,
                grid=both,
                line width=1pt
            ]
            \addplot[color=red, mark=square*] coordinates {(4,0.924) (8,0.933) (16,0.936) (32,0.938) (64,0.945)};
            \addplot[color=blue, mark=*] coordinates {(4,0.629) (8,0.650) (16,0.685) (32,0.704) (64,0.728)};
            \addplot[color=orange, mark=triangle*] coordinates {(4,0.622) (8,0.651) (16,0.665) (32,0.677) (64,0.684)};
            \end{axis}
        \end{tikzpicture}
    \end{minipage}\hfill
    % --- DISEASE (LOSS) ---
    \begin{minipage}[b]{0.24\textwidth}
        \centering
        \begin{tikzpicture}
            \begin{axis}[
                axis line style={line width=0.5pt},
                width=3.5cm,
                height=3.5cm,
                title={Disease},
                title style={yshift=-1ex},
                ylabel={Loss},
                xlabel={dim},
                xlabel shift={-3pt},
                ylabel shift={-3pt},
                xmode=log,
                log basis x={2},
                xtick={4, 8, 16, 32, 64},
                xticklabels={4, 8, 16, 32, 64},
                tick label style={font=\footnotesize},
                ymin=0.5, ymax=2.6,
                grid=both,
                line width=1pt
            ]
            \addplot[color=red, mark=square*] coordinates {(4,1.231) (8,0.973) (16,0.839) (32,0.867) (64,0.854)};
            \addplot[color=blue, mark=*] coordinates {(4,2.396) (8,1.360) (16,1.028) (32,1.021) (64,1.396)};
            \addplot[color=orange, mark=triangle*] coordinates {(4,2.254) (8,2.252) (16,2.253) (32,2.253) (64,2.251)};
            \end{axis}
        \end{tikzpicture}
    \end{minipage}\hfill
    % --- PHENOTYPE (LOSS) ---
    \begin{minipage}[b]{0.24\textwidth}
        \centering
        \begin{tikzpicture}
            \begin{axis}[
                axis line style={line width=0.5pt},
                width=3.5cm,
                height=3.5cm,
                title={Phenotype},
                title style={yshift=-1ex},
                xlabel={dim},
                xlabel shift={-3pt},
                xmode=log,
                log basis x={2},
                xtick={4, 8, 16, 32, 64},
                xticklabels={4, 8, 16, 32, 64},
                tick label style={font=\footnotesize},
                ymin=0.5, ymax=2.6,
                grid=both,
                line width=1pt
            ]
            \addplot[color=red, mark=square*] coordinates {(4,1.486) (8,1.313) (16,1.139) (32,0.903) (64,0.670)};
            \addplot[color=blue, mark=*] coordinates {(4,2.254) (8,2.254) (16,2.254) (32,2.254) (64,2.254)};
            \addplot[color=orange, mark=triangle*] coordinates {(4,2.254) (8,2.254) (16,2.253) (32,2.253) (64,2.251)};
            \end{axis}
        \end{tikzpicture}
    \end{minipage}
%    \vspace{-1ex}
    % --- LEGEND ---
    {\small
    \begin{tabular}{l@{\hspace{2em}}l@{\hspace{2em}}l}
        \textcolor{red}{---$\blacksquare$---} HGCN & 
        \textcolor{blue}{---$\bullet$---} GCN & 
        \textcolor{orange}{---$\blacktriangle$---} GAT
    \end{tabular}
    }
    
    \caption{\emph{ROC-AUC performance} and \emph{Loss} obtained by predictive models over hierarchical subgraphs at different embedding sizes. HGCN shows a markedly better performance when compared to Euclidean counterparts.}
    \label{fig:link_prediction_mixed}
\end{figure}
\paragraph{Results.}
The experimental results (see Figure~\ref{fig:link_prediction_mixed}) provide strong evidence for the advantage of the hyperbolic approach: HGCN consistently outperforms GCN and GAT, achieving ROC-AUC $> 0.90$ even with low-dimensional embeddings across both the Disease and HPO hierarchies. In contrast, the Euclidean models achieve at most a ROC-AUC of $0.8$, and only when using relatively high-dimensional embeddings on the Disease ontology.
Euclidean loss remains consistently higher, particularly in the Phenotype ontology, suggesting that the Euclidean models may make some incorrect predictions with high confidence, rather than achieving a better structural alignment; by contrast, the lower hyperbolic loss indicates more consistent confidence in the predicted hierarchy.
These results confirm a clear advantage of the hyperbolic approach on hierarchical structured data and motivate its exploration on larger, mixed-geometry environments. 
\subsection{Rank-Based Evaluation for Differential Diagnosis Simulation}
We simulated \emph{in silico} differential diagnosis as a KG completion task for the \texttt{HasDisease} relation over the targeted subgraph, benchmarking AttH against its Euclidean counterpart, AttE, and against HGCN. 
In this setting, models are trained to predict whether a \texttt{Person} node should be connected to a \texttt{Disease} node by a \texttt{HasDisease} edge.
To ensure a consistent evaluation across models, all methods were trained and tested using the same train--validation--test split. 
The test set was built by randomly holding out $10\%$ of the existing \texttt{HasDisease} edges as positive examples, which accounts for approximately $1\%$ of all edges in the targeted graph. All remaining positive edges in the graph, including the remaining \texttt{HasDisease} edges and all other relation types, were used for training and validation, accounting for approximately the $94.5\%$ and $4.5\%$ of all positive edges in the KG, respectively. The hyper-parameters of all the compared models were optimized by using the internal train/validation split to allow an unbiased, controlled evaluation.
\paragraph{Training of HGCN.}
HGCN was trained for binary link prediction using a binary cross-entropy loss over positive and negative edge samples. 
Given the positive training set, an equal number of negative samples was generated by edge corruption. 
The objective encourages observed edges to receive high predicted probabilities and corrupted edges to receive low predicted probabilities.
\paragraph{Training of shallow methods (AttE, AttH).}
Shallow methods were trained using a standard negative-sampling logistic loss (see equation \ref{eq:ATTH_loss}). 
Given a positive triple $(h,r,t)$, we sampled a fixed set of $50$ corrupted triples $(h,r,t')$ by replacing the true tail with entities drawn uniformly at random from the entire entity set, without any type or existence filtering (cf.\ Eq.~\ref{eq:ATTH_loss}). 
The training objective encourages the score of the positive triple to be higher than the scores of its negative counterparts, thereby placing the relation-transformed head representation close to the correct tail and far from sampled negative tails. 
AttE and AttH follow the same training principle, but differ in the geometry of the scoring function, as defined in Section~\ref{sec:background}-Par.~\ref{pg:shallow_Hmodels}. %\deleted[id=PM]{AttE computes distances in Euclidean space, whereas AttH computes relation-specific distances on the hyperbolic manifold.}
\paragraph{Inference.}
In inference, each test triple $(h,r,t)$ is treated as a query $(h,r,?)$, and candidate disease tails are ranked by their plausibility score. 
For shallow models, the plausibility of a candidate tail is defined by the score assigned to the corresponding triple, which depends on the distance between the candidate tail and the relation-transformed head. 
For HGCN, we simulate the same ranking setting by fixing a person node $h$ and scoring its link probability against every candidate disease node $t'$. 
For each person, candidate diseases are then ranked by decreasing predicted probability. 
Mean Rank (MR), Mean Reciprocal Rank (MRR), and H@10 were used to evaluate the quality of the ranked candidate diseases. For each test query, the model scores all candidate diseases and records the rank of the correct disease. MR: $\mathrm{MR} = \tfrac{1}{|\mathcal{Q}|}\sum_{q \in \mathcal{Q}} \mathrm{rank}_q,$
where $\mathcal{Q}$ is the set of test queries and $\mathrm{rank}_q$ is the position of the correct disease in the ranked list for query $q$. Lower MR values indicate better performance. MRR:
$\mathrm{MRR} = \tfrac{1}{|\mathcal{Q}|}\sum_{q \in \mathcal{Q}} \tfrac{1}{\mathrm{rank}_q}.$
A high MRR indicates that, on average, the correct disease appears near the top of the ranked list. Hits@10 is the proportion of test queries for which the correct entity is ranked among the top 10 candidate entities. Higher values indicate better link-prediction performance.

\begin{figure}[tbp]
    \centering
    \begin{tikzpicture}
    \begin{groupplot}[
        group style={group size=3 by 1, horizontal sep=0.95cm},
        axis line style={line width=0.5pt},
        width=4.6cm, height=3.9cm,
        xmode=log, xtick={2,4,10,26}, xticklabels={2,4,10,26},
        enlarge x limits=0.15, grid=major,
        every axis plot/.append style={line width=1pt},
        xlabel={Embedding dimension},
        label style={font=\footnotesize}, tick label style={font=\footnotesize},
        % metric name as a short HORIZONTAL tag above each panel (no rotated side labels)
        ylabel style={rotate=-90, at={(0.5,1)}, anchor=south, yshift=1pt, font=\footnotesize\bfseries},
    ]
    \nextgroupplot[ylabel={MRR},
        legend to name=grouplegend, legend columns=-1,
        legend style={draw=none, font=\footnotesize,
                      /tikz/every even column/.append style={column sep=10pt}}]
        \addplot[color=blue, solid, mark=*, mark size=1.6pt] coordinates {(2,0.142)(4,0.226)(10,0.508)(26,0.690)}; \addlegendentry{AttH}
        \addplot[color=yellow!90!black, solid, mark=square*, mark size=1.6pt] coordinates {(2,0.138)(4,0.160)(10,0.259)(26,0.543)}; \addlegendentry{AttE}
        \addplot[color=green!60!black, dashed, mark=diamond*, mark size=1.6pt] coordinates {(2,0.021)(4,0.085)(10,0.249)(26,0.397)}; \addlegendentry{HGCN}
    \nextgroupplot[ylabel={MR (log)}, ymode=log]
        \addplot[color=blue, solid, mark=*, mark size=1.6pt] coordinates {(2,128.34)(4,61.24)(10,24.72)(26,10.01)};
        \addplot[color=yellow!90!black, solid, mark=square*, mark size=1.6pt] coordinates {(2,136.90)(4,64.12)(10,38.16)(26,11.57)};
        \addplot[color=green!60!black, dashed, mark=diamond*, mark size=1.6pt] coordinates {(2,275.25)(4,97.48)(10,42.31)(26,23.82)};
    \nextgroupplot[ylabel={H@10}]
        \addplot[color=blue, solid, mark=*, mark size=1.6pt] coordinates {(2,0.295)(4,0.417)(10,0.733)(26,0.866)};
        \addplot[color=yellow!90!black, solid, mark=square*, mark size=1.6pt] coordinates {(2,0.288)(4,0.322)(10,0.464)(26,0.788)};
        \addplot[color=green!60!black, dashed, mark=diamond*, mark size=1.6pt] coordinates {(2,0.035)(4,0.196)(10,0.459)(26,0.635)};
    \end{groupplot}
    \end{tikzpicture}\\[2pt]
    \pgfplotslegendfromname{grouplegend}
    \caption{\emph{AttH, AttE, and HGCN comparison:} mean MRR, MR and H@10 for Disease-Only (RHS-only) prediction, for different embedding sizes. The average is computed across three independently seeded train--validation--test splits of the graph.}
    \label{fig:average-disease-rhs}
\end{figure}
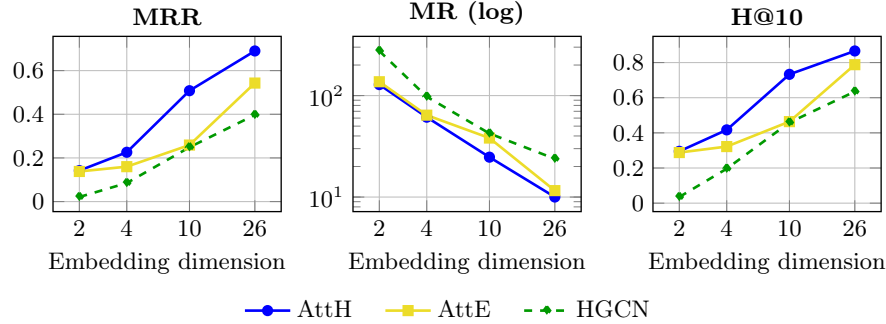

\paragraph{Results} Figure~\ref{fig:average-disease-rhs} shows that AttH outperforms AttE and HGCN across embedding dimensions on all three metrics (MRR, MR, and H@10), with best results at $d = 26$. %(MRR 0.690, MR 10.01, H@10 0.866 vs.\ AttE: MRR 0.543, MR 11.57, H@10 0.788; HGCN: MRR 0.397, MR 23.82, H@10 0.635). T
The exceptions are at the smallest dimensions: at $d = 2$ for MRR and H@10, and at $d \in \{2, 4\}$ for MR, where AttH shows no
consistent advantage over AttE (e.g., at $d = 2$: MRR 0.142 vs.\ 0.138, H@10 0.295 vs.\ 0.288). HGCN remains the weakest of the three across all dimensions and metrics, although its H@10 at $d = 10$ (0.459) nearly matches AttE (0.464).

% ============================================================
% FIGURE 2: Average - MRR, MR and H@10  Dis. (RHS-only)
% ============================================================

\section{Conclusion}

This preliminary study shows that hyperbolic GRL is promising for biomedical KGs combining ontology-derived hierarchies with transversal associations. HGCN outperformed Euclidean GNNs on isolated hierarchical subgraphs using substantially lower-dimensional embeddings, while AttH showed promising ranking performance in the rare-disease differential diagnosis task. These results suggest that hyperbolic embeddings can exploit biomedical hierarchical structure while remaining useful in heterogeneous patient-integrated graphs.

Future work will investigate hybrid models that combine Euclidean and hyperbolic components, including architectures with score-level fusion mechanisms that learn the relative contribution of Euclidean and hyperbolic distances to the final score, thereby better capturing both transversal and hierarchical relations.

\begin{credits}
\subsubsection{\ackname} This work was partially funded by Piano di Sviluppo di Ricerca (PSR2025) - Università degli Studi di Milano. 

Part of this work was funded by the National Plan for NRRP Complementary Investments (PNC) in the call for the funding of research initiatives for technologies and innovative trajectories in the health—project n. PNC0000003—AdvaNced Technologies for Human-centrEd Medicine (project acronym: ANTHEM).

Computational resources were provided by the INDACO Core facility (University of Milan).

\subsubsection{\discintname}
The authors have no competing interests to declare that are
relevant to the content of this article. 
\end{credits}

% {\small
% \putbib[USIINFMScThesis/biblioPaper]
% %\bibliographystyle{splncs04}
% %\bibliography{USIINFMScThesis/main_text_ordered_references} 
% }

% \end{bibunit}

\bibliographystyle{splncs04}
\bibliography{biblioPaper}

@book{hamilton2020graph,
  title = {Graph Representation Learning},
  DOI = {10.1007/978-3-031-01588-5},
  journal = {Synthesis Lectures on Artificial Intelligence and Machine Learning},
  publisher = {Springer International Publishing},
  author = {Hamilton,  William L.},
  year = {2020}
}

@article{callahan2020PheKnowlator,
  title="{An open source knowledge graph ecosystem for the life sciences.}",
  author="Callahan, TJ and others",
  journal={Scientific Data},
  volume=11,
  number=363,
  year={2024},
  publisher="{Nature Publishing}",
  doi="10.1038/s41597-024-03171-w"
}

@misc{HumanPhenotypeOntology,
  author       = {{Human Phenotype Ontology Consortium}},
  title        = {Human Phenotype Ontology},
  year         = {2025},
  howpublished = {\url{https://hpo.jax.org/}},
  note         = {Accessed: September 24, 2025}
}

@inproceedings{chami2019HGCN,
author = {Chami and others},
ignored={, Ines and Ying, Rex and Re, Christopher and Leskovec, Jure},
title = {Hyperbolic graph convolutional neural networks},
year = {2019},
booktitle = {Advances in Neural Information Processing Systems},
articleno = {438},
numpages = {12},
url = {https://dl.acm.org/doi/10.5555/3454287.3454725}
}

@inproceedings{nickel2017poincare,
  title={Poincar{\'e} embeddings for learning hierarchical representations},
  author={Nickel, Maximillian and Kiela, Douwe},
  booktitle={Advances in neural information processing systems},
  volume={30},
  year={2017},
  url={https://dl.acm.org/doi/10.5555/3295222.3295381}
}

@article{torgano2025rna,
    author = {Torgano, Francesco and others},
    ignored={Soto Gomez, Mauricio and Zignani, Matteo and Gliozzo, Jessica and Cavalleri, Emanuele and Mesiti, Marco and Casiraghi, Elena and Valentini, Giorgio},
    title = {RNA knowledge-graph analysis through homogeneous embedding methods},
    journal = {Bioinformatics Advances},
    volume = {5},
    number = {1},
    year = {2025},
    doi = {10.1093/bioadv/vbaf109},
}

@inproceedings{bendavid2002limitations,
  title={Limitations of learning via embeddings in Euclidean half-spaces},
  author={Ben-David, Shai and others},
  ignored={and Eiron, Nadav
and Simon, Hans Ulrich},
booktitle="Computational Learning Theory",
year="2001",
doi = "10.1007/3-540-44581-1_25"
}

@phdthesis{chami2021representation,
  author    = {Ines Chami},
  title     = {Representation learning and algorithms in hyperbolic spaces},
  school    = {Stanford University},
  year      = {2021},
  type      = {Ph.D. dissertation},
  url       = {https://searchworks.stanford.edu/view/13876470},
  note      = {Advisor: Christopher Ré}
}

@inproceedings{TransE,
 author = {Bordes, Antoine and others},
 ignored={Usunier, Nicolas and Garcia-Duran, Alberto and Weston, Jason and Yakhnenko, Oksana},
 booktitle = {Advances in Neural Information Processing Systems},
 title = {Translating Embeddings for Modeling Multi-relational Data},
 volume = {26},
 year = {2013},
 url = {https://dl.acm.org/doi/10.5555/2999792.2999923}
}

@inproceedings{TransH,
  title     = {Knowledge Graph Embedding by Translating on Hyperplanes},
  author    = {Zhen Wang and others},
  booktitle = {Proceedings of the Twenty-Eighth AAAI Conference on Artificial Intelligence},
  pages     = {1112--1119},
  year      = {2014},
  url       = {https://dl.acm.org/doi/10.5555/2893873.2894046},
  doi       = {10.5555/2893873.2894046}
}

@inproceedings{RotatE,
  title     = {RotatE: Knowledge Graph Embedding by Relational Rotation in Complex Space},
  author    = {Zhiqing Sun and others},
  booktitle = {International Conference on Learning Representations},
  year      = {2019},
  url       = {https://openreview.net/forum?id=HkgEQnRqYQ}
}

@article{Haendel20,
  title     = {How many rare diseases are there?},
  author    = {Haendel, M. and others},
  journal   = {{Nature Reviews Drug Discovery}},
  year      = {2020},
  volume    = {19},
  doi       = {10.1038/d41573-019-00180-y},
}

@inproceedings{RGCN,
author="Schlichtkrull, Michael and others",
ignored="
and Kipf, Thomas N.
and Bloem, Peter
and van den Berg, Rianne
and Titov, Ivan
and Welling, Max",
title="Modeling Relational Data with Graph Convolutional Networks",
booktitle="The Semantic Web",
year="2018",
doi="10.1007/978-3-319-93417-4_38"
}

@inproceedings{GCN,
  author       = {Thomas N. Kipf and
                  Max Welling},
  title        = {Semi-Supervised Classification with Graph Convolutional Networks},
  booktitle    = {Conference on Learning Representations},
  year         = {2017},
  url          = {https://arxiv.org/abs/1609.02907}
}

@inproceedings{GAT,
  author       = {Petar Velickovic and others},
  ignored      = {Guillem Cucurull and
                  Arantxa Casanova and
                  Adriana Romero and
                  Pietro Li{\`{o}} and
                  Yoshua Bengio},
  title        = {Graph Attention Networks},
  booktitle    = {International Conference on Learning Representations},
  year         = {2018},
  url       = {https://arxiv.org/abs/1710.10903}
}

@inproceedings{NBFnet,
author = {Zhu, Zhaocheng and others},
ignored={Zhang, Zuobai and Xhonneux, Louis-Pascal and Tang, Jian},
title = {Neural bellman-ford networks: a general graph neural network framework for link prediction},
year = {2021},
booktitle = {Advances in Neural Information Processing Systems},
articleno = {2256},
url ={https://dl.acm.org/doi/10.5555/3540261.3542517}
}

@inproceedings{MuRP,
    author = {Bala\v{z}evi\'{c} and others},
    IGNOREauthor = { Ivana and Allen, Carl and Hospedales, Timothy},
title = {Multi-relational poincar\'{e} graph embeddings},
year = {2019},
booktitle = {Advances in Neural Information Processing Systems},
articleno = {401},
numpages = {11},
url  = {https://dl.acm.org/doi/10.5555/3454287.3454688}
}

@inproceedings{AttH,
  title     = {Low-Dimensional Hyperbolic Knowledge Graph Embeddings},
  author    = {Ines Chami and others},
  ignored={and Adva Wolf and Da-Cheng Juan and Frederic Sala and Sujith},
  booktitle = {Annual Meeting of the Association for Computational Linguistics},
  ig_pages     = {6901--6914},
  year      = {2020},
  url       = {https://aclanthology.org/2020.acl-main.617/}
}

@inproceedings{HGNN,
author = {Liu, Qi and others},
ignored ={Nickel, Maximilian and Kiela, Douwe},
title = {Hyperbolic graph neural networks},
year = {2019},
booktitle = {Advances in Neural Information Processing Systems},
articleno = {739},
numpages = {12},
url = {https://dl.acm.org/doi/abs/10.5555/3454287.3455026}
}

@inproceedings{M2GNN,
author = {Wang, Shen and others},
ignored = {Wei, Xiaokai and Nogueira dos Santos, Cicero Nogueira and Wang, Zhiguo and Nallapati, Ramesh and Arnold, Andrew and Xiang, Bing and Yu, Philip S. and Cruz, Isabel F.},
title = {Mixed-Curvature Multi-Relational Graph Neural Network for Knowledge Graph Completion},
year = {2021},
doi = {10.1145/3442381.3450118},
booktitle = {Proceedings of the Web Conference 2021},
}

@inproceedings{ProductManifoldBIO,
  title={Product Manifold Representations for Learning on Biological Pathways},
  author={McNeela, Daniel and others},
  ignored={Sala, Frederic and Gitter, Anthony},
  booktitle={Great Lakes Bioinformatics Conference},
  year={2025},
  url= {https://arxiv.org/abs/2401.15478}
}

@article{HEM,
  title     = {Hyperbolic Hierarchical Knowledge Graph Embeddings for Biological Entities},
  author    = {Nan Li and others},
  ignored = {and Zhihao Yang and Yumeng Yang and Jian Wang and Hongfei Lin},
  journal   = {Journal of Biomedical Informatics},
  volume    = {147},
  pages     = {104503},
  year      = {2023},
  doi       = {10.1016/j.jbi.2023.104503},
}

@article{AlanisLobato2018,
  title     = {The latent geometry of the human protein interaction network},
  author    = {Alanis-Lobato, G. and others},
  journal   = {Bioinformatics},
  volume    = {34},
  number    = {16},
  year      = {2018},
  url       = {https://doi.org/10.1093/bioinformatics/bty206}
}

@article{GeOKG,
    author = {Jeong, Chang-Uk and others},
    ignored={Kim, Jaesik and Kim, Dokyoon and Sohn, Kyung-Ah},
    title = {{G}e{OKG}: geometry-aware knowledge graph embedding for Gene Ontology and genes},
    journal = {Bioinformatics},
    volume = {41},
    number = {4},
    year = {2025},
    month = {04},
    doi = {10.1093/bioinformatics/btaf160},
}

@article{HyperDrug,
  title = {Semi-supervised Hierarchical Drug Embedding in Hyperbolic Space},
  volume = {60},
  DOI = {10.1021/acs.jcim.0c00681},
  number = {12},
  journal = {Journal of Chemical Information and Modeling},
  author = {Yu,  Ke and others},
  ignored={Visweswaran,  Shyam and Batmanghelich,  Kayhan},
  year = {2020}
  }

@article{HyperMF,
  title     = {Hyperbolic matrix factorization improves prediction of drug-target associations},
  author    = {Aleksandar Poleksic},
  journal   = {Scientific Reports},
  volume    = {13},
  year      = {2023},
  doi       = {10.1038/s41598-023-27995-5}
}

@article{Phenomizer,
  title     = {Clinical Diagnostics in Human Genetics with Semantic Similarity Searches in Ontologies},
  author    = {Sebastian Köhler and others},
  journal   = {The American Journal of Human Genetics},
  volume    = {85},
  number    = {4},
  pages     = {457--464},
  year      = {2009},
  doi       = {10.1016/j.ajhg.2009.09.003}
}

@article{Phen2Disease,
  title     = {{Phen2Disease}: a phenotype-driven model for disease and gene prioritization by bidirectional maximum matching semantic similarities},
  author    = {Weiqi Zhai and others},
  journal   = {Briefings in Bioinformatics},
  volume    = {24},
  number    = {4},
  year      = {2023},
  url       = {https://doi.org/10.1093/bib/bbad172}
}

@article{Exomiser,
  title     = {Next-generation diagnostics and disease-gene discovery with the {Exomiser}},
  author    = {Damian Smedley and others},
  journal   = {Nature Protocols},
  volume    = {10},
  number    = {12},
  year      = {2015},
  doi       = {10.1038/nprot.2015.124}
}

@article{SHEPHERD,
  title     = {Few shot learning for phenotype-driven diagnosis of patients with rare genetic diseases},
  author    = {Emily Alsentzer and others},
  journal   = {NPJ Digital Medicine},
  volume    = {8},
  number    = {1},
  pages     = {380},
  year      = {2025},
  url       = {https://doi.org/10.1038/s41746-025-01749-1}
}

@article{chandak2023primekg,
  title     = {Building a knowledge graph to enable precision medicine},
  author    = {Payal Chandak and others},
  journal   = {Scientific Data},
  volume    = {10},
  number    = {1},
  pages     = {67},
  year      = {2023},
  url       = {https://doi.org/10.1038/s41597-023-01960-3}
}

@article{PhenopacketStore,
  title     = {A corpus of {GA4GH} phenopackets: {C}ase-level phenotyping for genomic diagnostics and discovery},
  author    = {Daniel Danis and others},
  journal   = {Human Genetics and Genomics Advances},
  volume    = {6},
  number    = {1},
   year      = {2025},
  doi       = {/10.1016/j.xhgg.2024.100371}
}

@inproceedings{Chen2024RareBenchSpecialists,
author = {Chen, Xuanzhong and others},
ignored={Mao, Xiaohao and Guo, Qihan and Wang, Lun and Zhang, Shuyang and Chen, Ting},
title = {RareBench: Can  {LLM}s Serve as Rare Diseases Specialists?},
year = {2024},
doi = {10.1145/3637528.3671576},
booktitle = {Conference on Knowledge Discovery and Data Mining},
}

@misc{naddeo2026microscope,
  title        = "{Hyperbolic Graph Neural Networks Under the Microscope: The Role of Geometry--Task Alignment}",
  author       = {Naddeo, Dionisia and Linkerh{\"a}gner, Jonas and Toschi, Nicola and Skenderi, Geri and Lachi, Veronica},
  year         = {2026},
  eprint       = {2602.01828},
  archivePrefix= {arXiv},
  primaryClass = {cs.LG},
  howpublished = {arXiv preprint arXiv:2602.01828}
}

@article{Ashburner2000,
  author    = {Ashburner, Michael and others},
  title     = {Gene Ontology: tool for the unification of biology},
  journal   = {Nature Genetics},
  year      = {2000},
  volume    = {25},
  number    = {1},
  pages     = {25--29},
  doi       = {10.1038/75556},
  pmid      = {10802651}
}

@article{Schriml2012,
  author    = {Schriml, Lynn Marie and others},
  title     = {Disease {Ontology}: a backbone for disease semantic integration},
  journal   = {Nucleic Acids Research},
  year      = {2012},
  volume    = {40},
  number    = {D1},
  doi       = {10.1093/nar/gkr972},
  pmid      = {22080554}
}

@article{Vasilevsky2022,
    author = {Vasilevsky, Nicole A and others},
    ignored = {Toro, Sabrina and Matentzoglu, Nicolas and Flack, Joseph E and Mullen, Kathleen R and Hegde, Harshad and Gehrke, Sarah and Whetzel, Patricia L and Shwetar, Yousif and Harris, Nomi L and Ngu, Mee S and Alyea, Gioconda L and Kane, Megan S and Roncaglia, Paola and Sid, Eric and Thaxton, Courtney L and Wood, Valerie and Abraham, Roshini S and Achatz, Maria Isabel and Ajuyah, Pamela and Amberger, Joanna S and Babb, Lawrence and Baker, Jasmine and Balhoff, James P and Berg, Jonathan S and Bhalla, Amol and Bofill-De Ros, Xavier and Braun, Ian R and Broeren, Eleanor C and Byer, Blake K and Byrne, Alicia B and Callahan, Tiffany J and Carmody, Leigh C and Chan, Lauren E and Clause, Amanda R and Cohen, Julie S and DeLuca, Marcello and Deuitch, Natalie T and Flowers, May and Fraser, Jamie and Fujiwara, Toyofumi and Gitau, Vanessa and Goldstein, Jennifer L and Gration, Dylan and Groza, Tudor and Gyori, Benjamin M and Hankey, William and Hilton, Jason A and Himmelstein, Daniel S and Hong, Stephanie S and Hoyt, Charles T and Huether, Robert and Hurwitz, Eric and Jacobsen, Julius O B and Kikuchi, Atsuo and Köhler, Sebastian and Korn, Daniel R and Lagorce, David and Laraway, Bryan J and Li, Jane Y and Malheiro, Adriana J and McLaughlin, James and Meldal, Birgit H M and Mohan, Shruthi and Moxon, Sierra A T and Munoz-Torres, Monica C and Nelson, Tristan H and Nicholas, Frank W and Ochoa, David and Olson, Daniel and Oprea, Tudor I and Oskotsky, Tomiko T and Osumi-Sutherland, David and Paris, Kelley and Parkinson, Helen E and Pendlington, Zoë M and Peng, Xiao P and Pizzino, Amy and Plon, Sharon E and Powell, Bradford C and Ratliff, Julie C and Rehm, Heidi L and Remennik, Lyubov and Riggs, Erin R and Roberts, Sean and Robinson, Peter N and Ross, Justyne E and Schaper, Kevin and Schilder, Brian M and Schmidt, Johanna L and Sharp, Elliott W and Similuk, Morgan N and Smedley, Damian and Sneddon, Tam P and Sparks, Rachel and Stefancsik, Ray and Stupp, Gregory S and Sundar, Shilpa and Takatsuki, Terue and Tammen, Imke and Tshering, Kezang C and Unni, Deepak R and Valasek, Eloise and Vanderver, Adeline and Wagner, Alex H and Webb, Ryan F and Welter, Danielle and Yaya-Stupp, Doron and Zankl, Andreas and Zhang, Xingmin Aaron and McMurry, Julie A and Chute, Christopher G and Hamosh, Ada and Mungall, Christopher J and Haendel, Melissa A, ClinGen DICER1 and miRNA-Processing Gene Variant Curation Expert Panel; ClinGen Hereditary Gene Curation Expert Panel; ClinGen Motile Ciliopathy Gene Curation Expert Panel; ClinGen Myeloid Malignancy Variant Curation Expert Panel; ClinGen TP53 Variant Curation Expert Panel; ClinGen X-Linked Inherited Retinal Disease Variant Curation Expert Panel},
    title = {Mondo: integrating disease terminology across communities},
    journal = {Genetics},
    volume = {232},
    number = {4},
    year = {2026},
    month = {04},
    doi = {10.1093/genetics/iyaf215},
}

\end{document}